%% file: rl.tex
\documentclass{article}

    \usepackage[preprint]{neurips_2022}

\usepackage[utf8]{inputenc} % allow utf-8 input
\usepackage[T1]{fontenc}    % use 8-bit T1 fonts
\usepackage{hyperref}       % hyperlinks
\usepackage{url}            % simple URL typesetting
\usepackage{booktabs}       % professional-quality tables
\usepackage{amsfonts}       % blackboard math symbols
\usepackage{nicefrac}       % compact symbols for 1/2, etc.
\usepackage{microtype}      % microtypography
\usepackage{xcolor}         % colors
\usepackage{amsmath,amssymb,amsmath,amsthm,amsfonts}
\usepackage{graphicx}
\usepackage{wrapfig}
\usepackage{subcaption}
\usepackage{caption}
\usepackage{multirow}
\usepackage{multicol}
\input{newcommands}

\title{Transferable Multi-Agent Reinforcement Learning with Dynamic Participating Agents}

\author{
Xuting Tang\\
Steven Institute of Technology\\
xtang18@stevens.edu\\
\And
Jia Xu\\
Steven Institute of Technology\\
jxu70@stevens.edu\\
\And
Shusen Wang\\
Xiaohongshu Inc\\
shusenwang@xiaohongshu.com
}

\begin{document}

\maketitle

\begin{abstract}
We study multi-agent reinforcement learning (MARL) with centralized training and decentralized execution.
During the training, new agents may join, and existing agents may unexpectedly leave the training.
In such situations, a standard deep MARL model must be trained again from scratch, which is very time-consuming.
To tackle this problem, we propose a special network architecture with a few-shot learning algorithm that allows the number of agents to vary during centralized training.
In particular, when a new agent joins the centralized training, our few-shot learning algorithm trains its policy network and value network using a small number of samples; when an agent leaves the training, the training process of the remaining agents is not affected.
Our experiments show that using the proposed network architecture and algorithm, model adaptation when new agents join can be 100+ times faster than the baseline. 
Our work is applicable to any setting, including cooperative, competitive, and mixed.
\end{abstract}

%%%%%%%%%%%%%%%%%%%%%%%%%%%%%%%%%%%%%%%%%%%%%%%%%%%%%%%%%%%%%%%%%%%%%%%%%%%%%%
%%%%%%%%%%%%%%%%%%%%%%%%%%%%%%%%%%%%%%%%%%%%%%%%%%%%%%%%%%%%%%%%%%%%%%%%%%%%%%
%%%%%%%%%%%%%%%%%%%%%%%%%%%%%%%%%%%%%%%%%%%%%%%%%%%%%%%%%%%%%%%%%%%%%%%%%%%%%%
%%%%%%%%%%%%%%%%%%%%%%%%%%%%%%%%%%%%%%%%%%%%%%%%%%%%%%%%%%%%%%%%%%%%%%%%%%%%%%

\section{Introduction}
\label{sec:intro}

This paper studies Multi-Agent Reinforcement Learning (MARL) under the setting of Centralized Training and Decentralized Execution (CTDE).
Existing work such as MADDPG~\cite{lowe2017multi}, COMA~\cite{foerster2018counterfactual}, and QMIX~\cite{pmlr-v80-rashid18aQMIX} have demonstrated strong empirical performance.
Independent RL \cite{tan1993multi,Tampuu_2017DQN_IQL,foerster2017stabilising} and fully-centralized RL~\cite{lazaridou2016multiagent,sukhbaatar2016learning} are alternatives to CTDE but they are less practical.

% Independently RL (IRL) and fully-centralized RL are alternatives to but less practical than CTDE.
% IRL is a straightforward extension of single-agent RL to MARL \cite{tan1993multi,Tampuu_2017DQN_IQL,foerster2017stabilising}.
% Despite its simplicity, the performance of IRL is typically worse than CTDE due to the non-stationary environment issue~\cite{foerster2018counterfactual,lowe2017multi}. 
% Fully-centralized RL performs well but is not scalable because the dimensions of action space and observation space grow exponentially with the number of agents~\cite{lazaridou2016multiagent,sukhbaatar2016learning}. 

% Independently learning agents is the most straightforward approach to extend single agent RL algorithms to multi-agent settings~\cite{Tampuu_2017DQN_IQL,foerster2017stabilising}, but this approach performs poorly because of the non-stationary environment issue~\cite{foerster2018counterfactual,lowe2017multi}. 
% Fully centralized approaches~\cite{lazaridou2016multiagent,sukhbaatar2016learning} can avoid the non-stationary environment issue but it has scalability issues as the dimensions of action space and observation space grow exponentially with the number of agents. 
% To alleviate scalability issues and the difficulties caused by non-stationary environment, many centralized training and decentralized execution algorithms were proposed, such as MADDPG~\cite{lowe2017multi}, COMA~\cite{foerster2018counterfactual}, and QMIX~\cite{pmlr-v80-rashid18aQMIX}.

%---------------------------------Figure---------------------------------%
\begin{wrapfigure}[17]{r}{0.55\textwidth}
\vspace{-1mm}
  \begin{center}
    \includegraphics[width=0.48\textwidth]{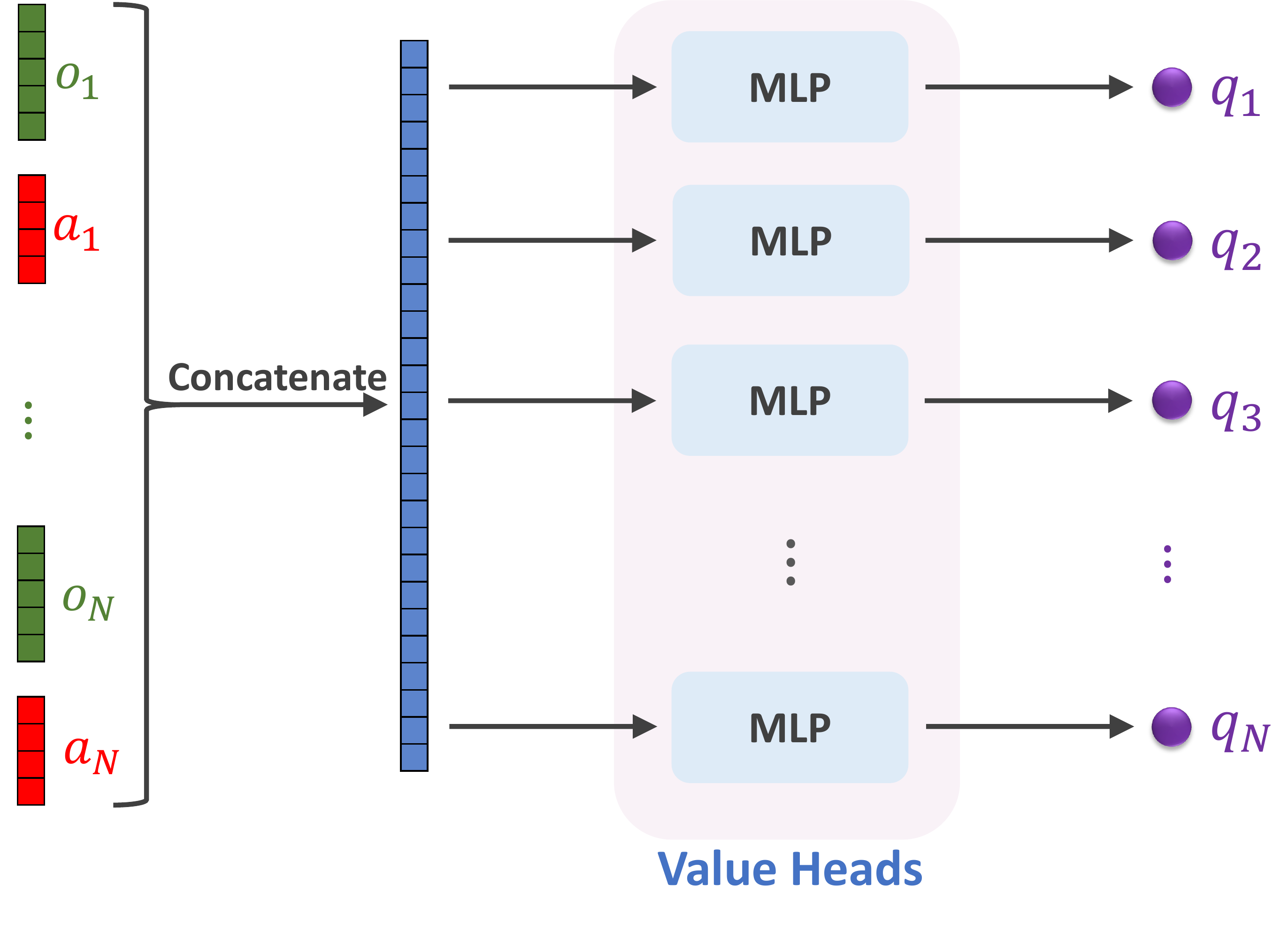}
    \vspace{-1mm}
  \end{center}
  \caption{Architecture of the value networks (critics) of MADDPG.}
  \label{fig:critic_old}
  \vspace{-1mm}
\end{wrapfigure}
%---------------------------------Figure---------------------------------%

CTDE assumes that a collection of $N$ agents can partially observe the environment and independently make decisions.
During training, each agent only communicates with the central coordinator to update its policy.
The centralized training of MADDPG, COMA, and QMIX has a limitation that the number of agents must be fixed during training.
New agents are not allowed to participate, and existing agents are not allowed to drop halfway during training.
We show in Figure~\ref{fig:critic_old} the centralized value network (critic) of MADDPG.
On the one hand, if the number of agents changes, the input shape of the network will change, and a new model has to be trained all over again from scratch.
On the other hand, if a new agent joins, its value and policy networks have not been learned yet. Such an agent will slow down the centralized training.
% \red{Explain why number of agents may change, provide some real life scenarios.}

% A major shortcoming of existing MARL frameworks is that they does not allow the participation of new agents nor agents drop from training. 
% If a new agent participates halfway or an agent drops during the centralized training, the input size of the networks would change. Take MADDPG framework as an example, its value network takes as input the concatenation of observations and actions from all agents, as shown in Figure~\ref{fig:critic_old}. The participation of a new agent will result in a larger input dimension and a sudden absence of an agent will shrink the input dimension; policy network has the same situation, as it takes as input the observations of all agents, the size of the observation is determined by the number of agents in the system, its input size will differ too. 
% Consequently, the trained model has to be discarded and a new model has to be trained from scratch. 

\begin{figure}
\centering
%  \begin{subfigure}{0.9\textwidth}
  \centering
    \includegraphics[width=\linewidth]{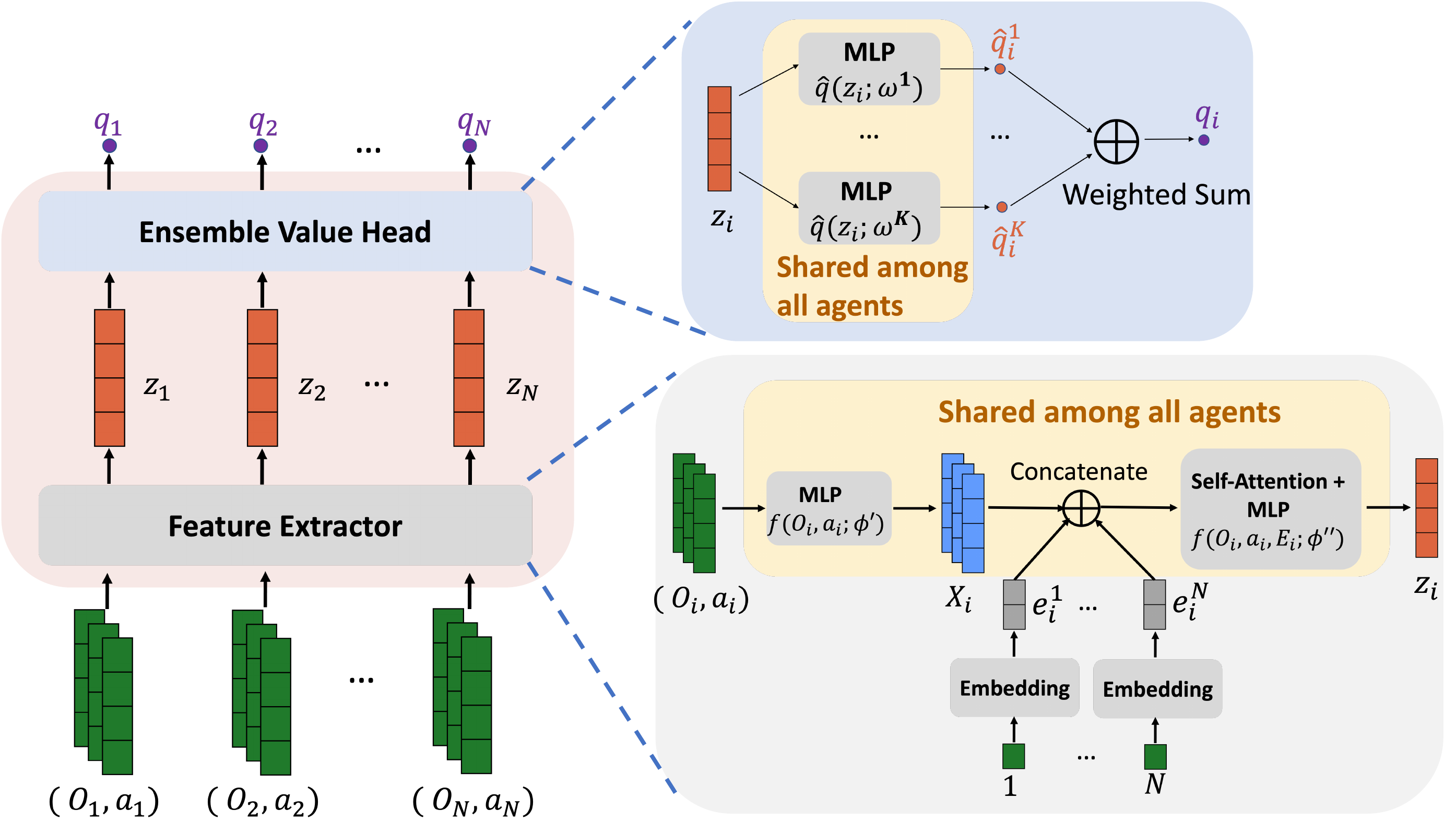}
    \caption{Architecture of the proposed value network.}
    \label{fig:value}
%  \end{subfigure}
  %
%   \begin{subfigure}{0.9\textwidth}
%   \centering
%     \includegraphics[width=0.35\linewidth]{policy_sim_whole.png}
%     \caption{Policy network.}
%     \label{fig:policy}
%   \end{subfigure}
%  \caption{Architecture of the proposed value network}
%  \label{fig:whole_structure}
\end{figure}

This paper addresses the problem of dynamic participating agents during centralized training.
For simplicity, our work is built upon MADDPG.
Our work can be extended to other multi-agent actor-critic methods such as COMA.
Our novelties include the new model architecture shown in Figures~\ref{fig:value} and \ref{fig:policy} and a few-shot learning algorithm.
\begin{itemize}
    \item 
    The proposed model architecture uses self-attention~\cite{bahdanau2014neural,cheng2016long,vaswani2017attention} as a feature extractor for allowing the number of agents to vary.
    Upon the self-attention feature extractor, the network has specially designed heads for reducing the number of parameters to learn when new agents participate.
    \item
    We propose a few-shot learning algorithm for enabling newly joined agents to quickly get adapted.
    The few-shot learning transfers knowledge from existing agents to newly joined agents.
    Our experiments show that the few-shot learning algorithm makes the training of newly joined agents hundreds of times faster than the baseline.
\end{itemize}

The rest of this paper is organized as follows.
Section~\ref{sec:related} discusses related work. Section~\ref{sec:background} introduces some background information of MARL. 
Section~\ref{sec:model_arch} describes the model architecture, followed by the detailed description of our few-shot learning algorithm in Section~\ref{sec:few-shot algo}. Section~\ref{sec:experiment} presents empirical results demonstrating the effectiveness and scalability of our approach.

% To address the problem of dynamic participating agents during training, we propose to use self-attention in network structure. Apart from allowing dynamic number of participating agents, the proposed network structure is designed to enables quick adaption of newly join agents, the adaption procedure will be introduced in Section~\ref{sec:few-shot}.
% Our approach is an actor-critic based method, Figure ~\ref{fig:whole_structure} shows the architecture of the proposed value network; policy network has similar design and it is omitted for simplicity.
% To enable quick adjustment and adaptation, the learned knowledge needs to be transferable, we achieve this by sharing common network components and reducing trainable parameters during the quick adaptation. When a new agent participates, we can reuse the majority parts of the network, instead of training the whole model, we use few-shot learning to train a few parameters, then the whole model will be in good shape. On the other hand, if an agent drops from the training, because of the network flexibility thanks to attention mechanism, the remaining agents can be unaffected and their training process stays the same.

%---------------------------------Figure---------------------------------%
% \begin{figure*}[!t]
% 	\begin{center}
% 		\includegraphics[width=0.9\textwidth]{whole_structure.png}
% 	\end{center}
% 	\caption{The proposed network for centralized training. For simplicity, only the value networks (critics) is shown.
% 	}
% 	\label{fig:whole_structure}
% 	\vspace{-2mm}
% \end{figure*}

%---------------------------------Figure---------------------------------%

% \paragraph{Problem 1:} What if a participating agent drops or a new agents participate during the training?

% \paragraph{Problem 2:} After training, if the number of agents changes, can the new agent be trained in a few shots?

% \paragraph{Problem 3:} How to protect privacy?

% This work has the following contributions:
% \begin{itemize}
%     \item 
%     We study the problem of dynamic participation during the training of MARL.
%     Note that the prior MARL frameworks assume fixed number of agents.
%     \item
%     We design a flexible network structure that allows existing agents to drop and new agents to join.
%     There are three key ingredients in the design:
%     1) we use attention layers to handle varying number of agents;
%     2) to address the problem of cold start, we design an ensemble value head and an ensemble policy head that can be trained in a few shots;
%     3) to ease few-shot learning, special agent embeddings are added.
% \end{itemize}

\section{Related Work} \label{sec:related}

Many MARL methods have been developed in recent years.
MADDPG \cite{lowe2017multi}, COMA \cite{foerster2018counterfactual}, and QMIX \cite{pmlr-v80-rashid18aQMIX} are popular MARL methods that use centralized training and decentralized execution (CTDE).
MADDPG is applicable to mixed cooperative-competitive settings, whereas COMA and QMIX are limited to fully cooperative settings.
MADDPG and COMA are actor-critic methods, while QMIX is a value-based method.
% MADDPG studies MARL in settings where agents can have arbitrary reward structures, it is a actor-critic-based method and it allows each agent to have its own critic and policy; COMA, another actor-critic-based approach, focuses on fully cooperative setting where all the agents have the same reward and thereby the same objective function, each agent has its private policy and a centralized critic is shared among all agents; QMIX works under fully cooperative setting, it is a value-based method which decomposes the global value function into monotonically constrained per-agent value function.

More recent papers---MAAC \cite{iqbal2019actor}, REFIL \cite{iqbal2021randomized}, COPA \cite{liu2021coach}, and UPDeT \cite{hu2020updet}---are more closely related to our work.
Our work and theirs all apply attention to MARL.
Nevertheless, our work has substantial differences from theirs.
MAAC is an actor-critic method for mixed cooperative-competitive settings.
Different from our motivation, MAAC uses attention for improving agents' performance.
Similar to QMIX, REFIL and COPA are value-based methods for the fully-cooperative setting.
Trained simultaneously on tasks with different scenarios,  policies learned by REFIL can deal with varying types and quantities of agents during execution.
COPA does not strictly follow CTDE: it has a coach that, during training, periodically distributes global information to the agents.
The aforementioned methods do not solve the problem where agents or entities dynamically join or leave the training process which is the focus of our work.

\section{Background} \label{sec:background}

\paragraph{Markov Games.}
Most of the recent works, e.g., MADDPG and MAAC, focus on partially observable Markov game \cite{littman1994markov} which is a multi-agent extension of Markov decision processes (MDPs).
A Markov game with $N$ agents consists of a set of states $\SM$, %which describes all possible configurations of all agents, 
a set of actions $\AM_1, \dots, \AM_N$, 
a set of observations $\OM_1, \dots, \OM_N$, %for the $N$ agents, 
a state transition function $\TM$                                                                                                           : $\SM \times \AM_1 \times \dots \times \AM_N \rightarrow \SM$, %which produces next state given the current state and actions of all agents, 
and a reward function $r_i                                                                                                                  : \SM \times \AM_i \rightarrow \RB$ for each agent. %, which is a function of the state and the agent's action, $r_i: \SM \times \AM_i \rightarrow \RB$. 
Each agent receives a private observation $\OM_i$ which is partial information of the global state $S$. Each agent learns its policy $\pii_i: \OM_i \rightarrow P(\AM_i)$ that maps its observation to a distribution over a set of actions. 
The goal of each agent $i$ is to maximize its own total expected return which is a weighted sum of its future rewards.
Our work is a further extension of the partially observable Markov game where $N$ may vary during the training. 
If $N$ remains the same, our setting is the same as the Markov game.
When $N$ changes, a new Markov game gets started.
We assume all agents share the same action space.

\paragraph{Actor-Critic.}
This work focuses on the actor-critic method \cite{barto1983neuronlike} for MARL.
Each agent has a policy network (actor) that can be either stochastic \cite{foerster2018counterfactual} or deterministic \cite{lowe2017multi}.
During the centralized training, the central coordinator maintains $N$ value networks (critics) that evaluate the state $S$ and actions $A_1, \cdots , A_N$.
Under the fully cooperative setting, the $N$ critics are the same because all the agents receive the same reward \cite{foerster2018counterfactual}.
Under the competitive or mixed cooperative-competitive setting, there are $N$ different critics \cite{lowe2017multi}.

\paragraph{MADDPG and MATD3.}
MADDPG \cite{lowe2017multi} is an actor-critic method for MARL.
During the centralized training, MADDPG learns a critic and a policy for each agent $i$.
% It learns a centralized critic $Q_i$ and a decentralized policy $\pi_i$ for each agent $i$, where $Q_i$ takes as input the actions and observations of all agents and $\pi_i$ takes as input the local observation of agent $i$. 
MATD3~\cite{ackermann2019reducing} applies the Twin Delayed Deep Deterministic policy gradient algorithm~\cite{fujimoto2018addressing} (TD3) to stabilize the training of MADDPG.
TD3 helps address function approximation error in actor-critic methods. 
It uses Clipped Double Q-learning to alleviate the overestimation problem and adds random noise to the target policy as a smoothing regularization strategy. 
To further stabilize training, it updates the policy network at a lower frequency than the value network.

%---------------------------------Figure---------------------------------%
\begin{wrapfigure}[17]{r}{0.45\textwidth}
\vspace{-15mm}
  \begin{center}
    \includegraphics[width=0.35\textwidth]{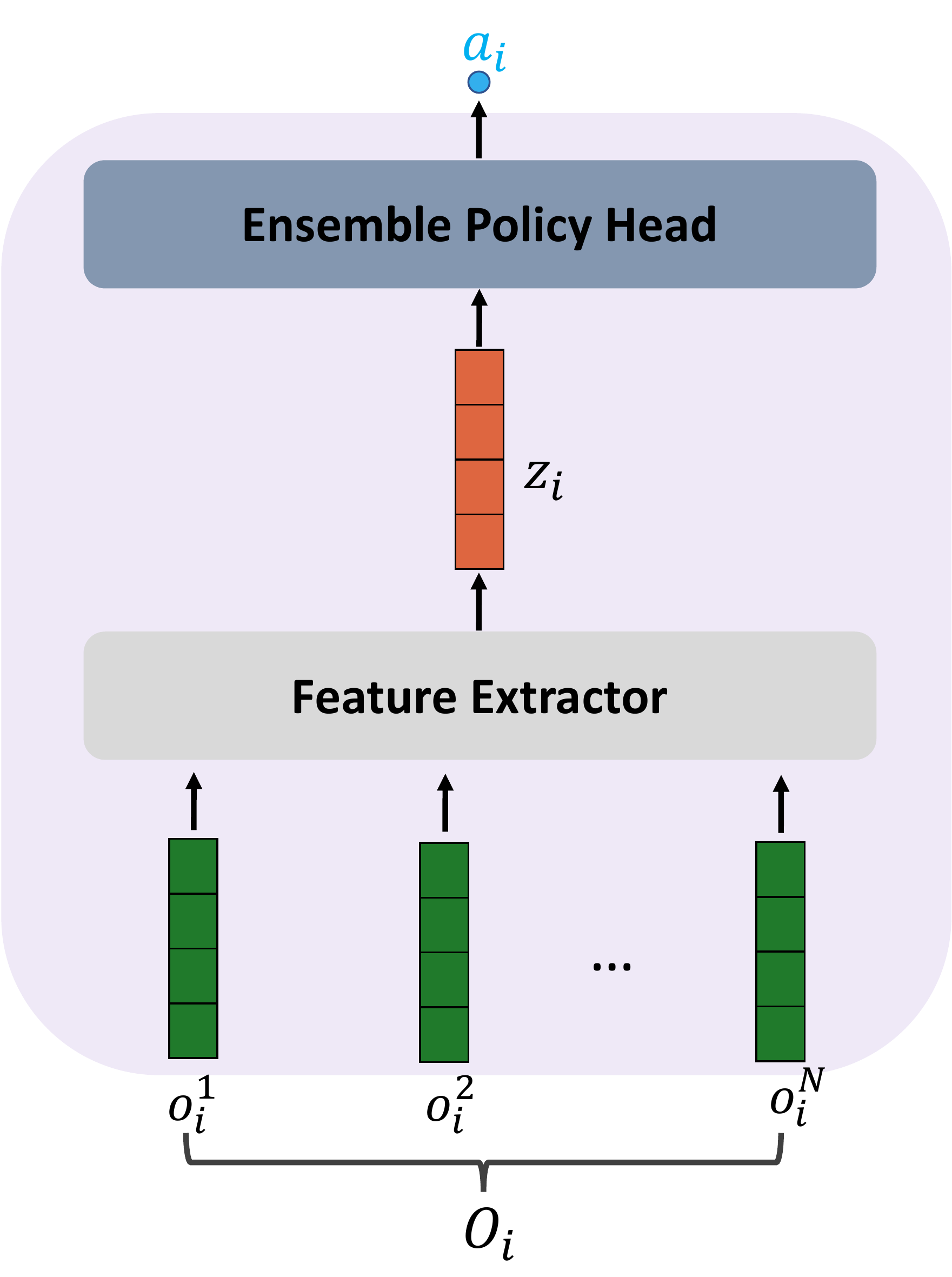}
    \vspace{-3mm}
  \end{center}
  \caption{Architecture of the proposed policy network.}
  \label{fig:policy}
  \vspace{15mm}
\end{wrapfigure}
%---------------------------------Figure---------------------------------%

\section{Model Architecture} \label{sec:model_arch}
Here we introduce our novel model architecture that can handle dynamic participating agents during centralized training. 
When new agents join the training, the model can quickly adapt to this change in a few shots; when agents drop from the training, the training process of the remaining agents is not affected. %If there are more (or less) agents in testing, the network just needs to be trained for a few shots before it can be applied to the new environment.
The proposed value network structure is shown in Figure \ref{fig:value}.
It is composed of a feature extractor and an ensemble head.
The former is for handling dynamic participating agents, and the latter is for quick adaptation of newly joined agents. 
The policy network shown in Figure~\ref{fig:policy} has the same structure, so we do not further discuss it.
% Since value network and policy network have similar structure, we will mainly focus on explaining the value network structure for simplicity and clarity.

Assume there are $N$ agents in the environment. 
Each agent has a value network and a policy network.
Each value network takes as input all the observations and all the actions of the $N$ agents.
Each policy network takes as input the local observation of the corresponding agent. 
Generally, agent $i$'s observation consists of information from all agents that is observable to agent $i$ (e.g., the relative positions of all agents with respect to agent $i$), it can be denoted as $O_i = [o^1_i, \cdots, o^N_i ]$, where $o^j_i$ represents the status of agent $j$ that is observable to agent $i$. Note that $O_i$ is the local observation of agent $i$.
When one more agent joins the training, $O_i$ becomes $[o^1_i, \cdots, o^{N+1}_i ]$; if the last agent drops from the training, $O_i$ becomes $[o^1_i, \cdots, o^{N-1}_i ]$.

%Instead of taking the concatenation of all observations and actions as input, 
Our proposed value network takes as input the observation, $O_i$, and the action, $a_i$, of each agent, as shown in Figure \ref{fig:value}. 
The inputs, $(O_1, a_1), \cdots, (O_N, a_n)$, are first processed by the feature extractors, and then the extracted features $[z_1, \cdots, z_N]$ become the input of the ensemble value heads.
The value heads output the $N$ values, $[q_1, \cdots, q_N] $. 
As shown in Figure~\ref{fig:policy}, policy network $i$ takes as input the observation $O_i$ from agent $i$, and outputs an action (or a probability distribution over the actor space) for agent $i$.

% When an agents dynamically participate or drop, the past experience becomes useless.
% Therefore, we study only on-policy methods without using experience replay.
% Arguably, experience replay can be useful if agents participate or drop infrequently.
% Our framework can be extended to off-policy control methods, e.g., DDPG, SAC, TD3; we leave the off-policy control to our future work.

\subsection{Feature Extractors}

A feature extractor is composed of fully connected feed-forward layers (MLPs), a multi-head self-attention layer, and an agent embedding layer. 
%Let $z_i = f (O_i, a_i, E_i; \phi )$ be the feature extractor in agent $i$'s value network (or $z_i = f (O_i, E_i; \phi )$ for policy feature extractor), where $E_i = [e^1_i, \cdots, e^N_i]$ is agent embeddings of all agents from agent $i$'s perspective, we will explain this in detail shortly. 
Each agent has a value network. 
The $N$ value networks share a common feature extractor, and so are the $N$ policy networks.

% Given the objective function $L$ and the gradient $\nabla_{z_i} L$, we can update the parameters with learning rate $\eta^\phi$ by 
% \begin{equation}
%     \phi 
%     \; \leftarrow \;
%     \phi \, \pm \, \eta^\phi \cdot \sum_{i=1}^N \big( \nabla_{\phi} z_i \big) \cdot \big( \nabla_{z_i} L \big) .
% \end{equation}
% Here, the sign $\pm$ depends on whether it is updating value feature extractor (gradient descent) or policy feature extractor (gradient ascent).

\paragraph{Agent ID Embedding.}
% In order to help self-attention layer better identify which agents it should pay more attention to, we add agent embeddings.
% In our setting, agents can be heterogeneous and there can be $K$ different types of agents, with $K \leq N$. For example, in predator-prey task, there are 2 types of agents, if the prey knows which agents are predators, it can better escape; if the predators know which is the prey, they can better surround it and catch it. 
The agent ID embedding layer maps an agent ID to a vector.
It plays a similar role as the positional encoding in Transformer~\cite{vaswani2017attention}.
Without the embedding, exchanging $(O_2, a_2)$ and $(O_3, a_3)$ will not affect $q_1$.
It means that without the embedding, the first agent's value head cannot distinguish the second and third agents.
This can be problematic in practice, for example, a predator may chase another predator instead of the prey.
% The goal of having agent embeddings is to help attention layer better identify which agents it should pay more attention to. 
% Every agent has an unshared agent embedding layer which consists of $N$ blocks, each corresponds to the embedding of an agent. 
% The embedding layer is defined by:
% \begin{equation}
%     e^j_{i} = \epsilon^j_{i} \cdot s_j, \qquad \forall i,j = 1, \cdots , N,
% \end{equation}
% where $\epsilon^j_{i} \in \mathbb{R}^{K}$ is the parameter of the $j$-th embedding block of the $i$-th agent's embedding layer and $s_j$ is the ID of agent $j$. 
As shown in Figure \ref{fig:value}, an MLP takes as input $(O_i, a_i) $ and outputs $X_i$. %$X_i = [ x^1_i, \cdots , x^N_i ]$.
Then the agent embedding vector $e_{i}^j$ is concatenated to $x_{i}^j$ before being fed to the attention layer.

\paragraph{Multi-Head Self-Attention Layer.}
To handle varying numbers of agents during training time, we apply multi-head self-attention to our model structure. A self-attention layer does not require the input to have a fixed length. For example, in NLP, a sequence-to-sequence model with attention can handle sentences of varying lengths. Therefore, with self-attention, our model can handle observations and actions from varying numbers of agents.

\subsection{Ensemble Heads}
While agents can use independent MLPs (without parameter sharing) as their value heads and policy heads, we take a different approach for the sake of knowledge transfer. %Figure~\ref{fig:whole_structure} shows the details of the ensemble heads.

\paragraph{Ensemble Value Heads.} 

The head of each value network is an ensemble of $K$ MLPs, $\widehat{q} (\cdot ; \omega^1) , \cdots , \widehat{q} (\cdot ; \omega^K)$. The $K$ MLPs are shared among all agents and are referred to as the candidate value blocks.
Here, $K$ ($\leq N$) is a hyper-parameter; if we know the number of types of agents as the prior knowledge, we can set $K$ to be the number of types.
% As mentioned before, $K$ is the number of agent types.
Each of the value heads is a weighted average of the $K$ candidate value blocks.
Agent $i$'s action-value function is approximated by
\begin{equation}
    q_i
    \; = \; \sum_{k=1}^K \alpha_i^k \widehat{q} \big( z_i ; \omega^k \big) , \qquad
    \textrm{where} \;\; \big[ \alpha_i^1, \cdots , \alpha_i^K \big] \; = \; \mathop{\rm softmax} \Big( \psi_i^1, \cdots , \psi_i^K  \Big) .
\end{equation}
% The value head of the agent $i$ takes as input $z_i$.
% Note that $q_i$ depends on all of $o^1, \cdots , o^N$ because $z_i$ depends on all of $o^1, \cdots , o^N$.

The parameters of the value heads, $\omega = (\omega^1, \cdots , \omega^k)$ and $\psi_i = (\psi_i^1, \cdots , \psi_i^K)$, can be updated by temporal difference (TD) algorithm. Here, we refer to $\psi_i$ as agent $i$'s selector of the candidate value blocks.

% Let $[(O_1, a_1), \cdots, (O_N, a_N)]$ and $[(O_1' a_1'), \cdots, (O_N', a_N')]$ be the observations and actions in two consecutive iterations;
% let $r_i$ be the reward, along with $O_i'$ and $a_i'$, given by the environment to agent $i$.
% Let $y_i = r_i + \gamma \cdot q_i'$ be the TD target and $\delta_i = q_i - y_i$ be the TD error.
% Let $L_i = \frac{1}{2} \delta_i^2 $ be the loss.
% Parameters $w$ and $\psi$ can be updated by
% \begin{equation}
% \begin{split}
%     w^k &\leftarrow  w^k - \eta^w \cdot \sum_{i=1}^N \delta_i \cdot \alpha_i^k \cdot \nabla_{w} \widehat{q} \big( z_i ; w^k \big),\\
%     \psi^k_i &\leftarrow \psi^k_i - \eta^\psi \cdot \delta_i \cdot \widehat{q} \big( z_i ; w^k \big) \cdot \nabla_{\psi^k_i}\alpha^k_i
% \end{split}
% \end{equation}
% where $\eta^w$ and $\eta^\psi$ are learning rates.

% In addition, compute
% \begin{equation}
%     \nabla_{z_i} L_i
%     \; = \;
%     \delta_i  \cdot \sum_{k=1}^K \alpha_i^k \cdot \nabla_{z_i} \widehat{q} \big( z_i ; w^k \big) 
% \end{equation}
% to continue the backpropagation.

\paragraph{Ensemble Policy Heads.}

An ensemble policy head maps extracted feature $z_i$ to an action or a probability distribution over the action space. The ensemble policy head is a weighted average of $K$ shared MLPs.
For a non-deterministic policy $ \widehat{\pi} = \pi (\cdot \, | z_i, \theta)$, the $d$-dim vector $\widehat{\pi}$ represents a discrete distribution over the action space $\AM_i$. 
Let $\widehat{\pi} ( \cdot ; \theta^1), \cdots , \widehat{\pi} (\cdot ; \theta^K )$ be $K$ candidate policy blocks. The policy head of the $i$-th agent is
\begin{equation}
\pi_i = \sum_{k=1}^K \beta_i^k \cdot \widehat{\pi} \big( z_i ; \theta^k \big), \quad \text{where} \big[ \beta_i^1 , \cdots , \beta_i^K \big] = \mathop{\rm softmax} \Big( \zeta_i^1 , \cdots , \zeta_i^K \Big).
\end{equation}
Similarly, we refer to $\zeta_i = (\zeta_i^1 , \cdots , \zeta_i^K)$ as agent $i$'s selector of candidate policy blocks. 
For a deterministic policy, it follows the same network construction, so it is omitted for simplicity.

\section{Few-Shot Learning for Dynamic Participants} \label{sec:few-shot algo}
\label{sec:few-shot}

% During training, some of the $N$ agents might drop from the training, or new agents can also join the training.
The proposed framework can trivially handle dropping agents without doing anything; it just ignores the agents that do not respond.
But the participation of a new agent is nontrivial to handle.
When a new agent (let it be the $(N+1)$th) participates in the training, if it uses an independent value network and policy network, training them from random initialization would require a large number of samples, which poses a challenge of ``cold start''.
% If an agent used independent networks as its value network and policy network, training these new networks from random initialization would require a large number of samples.
% Before they were trained, the new value network could not provide supervision to the new policy network, and the new policy network could not effectively explore to improve the new value network. 
Our design of the network structure is for the purpose of getting the new agents trained in a few shots.
In this section, we will describe the trainable parameters and the training procedure using few-shot learning.

% Table~\ref{tab:share} summarizes independent and shared components of value networks and policy networks.
The shared components in the networks are relatively well trained and ready for use, thus they can be fixed during the few-shot learning. The new selectors and agent ID embeddings are unshared and agent-specific, so they will be the trainable parameters during the few-shot learning. 
%---------------------------------Table---------------------------------%
% \begin{table}[]
%     \caption{Summarization of the independent and shared components in value networks and policy networks. Note that value networks and policy networks have different sets of shared parameters.}
%     \begin{center}
%     \begin{tabular}{c|ccc|cc}\toprule
%               & \multicolumn{3}{c|}{Feature Extractor} & \multicolumn{2}{c}{Ensemble Head}\\
%               & MLP & Agent Embeddings & Self-attention & $K$ Candidate Blocks      & Selector \\\midrule
%         Shared & Yes & No               & Yes            & Yes          & No     \\\bottomrule
%     \end{tabular}
%     \end{center}
%     \label{tab:share}
% \end{table}
%---------------------------------Table---------------------------------%
The new value head and policy head are weighted averages of the candidate blocks:
\begin{equation}
\begin{split}
q_{N+1} & = \sum_{k=1}^K \alpha^k_{N+1} \cdot \widehat{q} (z_{N+1} ; \omega^k), \quad \text{where} \big[ \alpha^1_{N+1}, \cdots , \alpha^K_{N+1} \big] = \mathop{\rm softmax} \Big( \psi^1_{N+1}, \cdots , \psi^K_{N+1}  \Big),\\
\pi_{N+1} & = \sum_{k=1}^K \beta^k_{N+1} \cdot \widehat{\pi} (z_{N+1} ; \theta^k), \quad \text{where} \big[ \beta^1_{N+1}, \cdots , \beta^K_{N+1} \big] = \mathop{\rm softmax} \Big( \zeta^1_{N+1}, \cdots , \zeta^K_{N+1}  \Big).
\end{split}
\end{equation}
There are $2K$ trainable scalar parameters for the selectors: $\psi^1_{N+1}, \cdots, \psi^K_{N+1}$, and $\zeta^1_{N+1}, \cdots , \zeta^K_{N+1}$.
Additionally, the agent ID embedding for the new agent also needs to be trained during the few-shot learning. 
Before finishing the few-shot learning, we do not update the shared network parameters; after the few-shot learning is done, we resume the regular training process for all agents.
Compared to using independent networks, our approach has significantly fewer parameters to train for the new agent.

\section{Experiments} \label{sec:experiment}

In this section, we evaluate the effectiveness and scalability of our method in different environments. 
We further conduct ablation studies to analyze the importance of each network component.

\subsection{Environments}
\begin{figure}
\centering
  \begin{subfigure}{0.4\textwidth}
  \centering
    \includegraphics[width=0.7\linewidth]{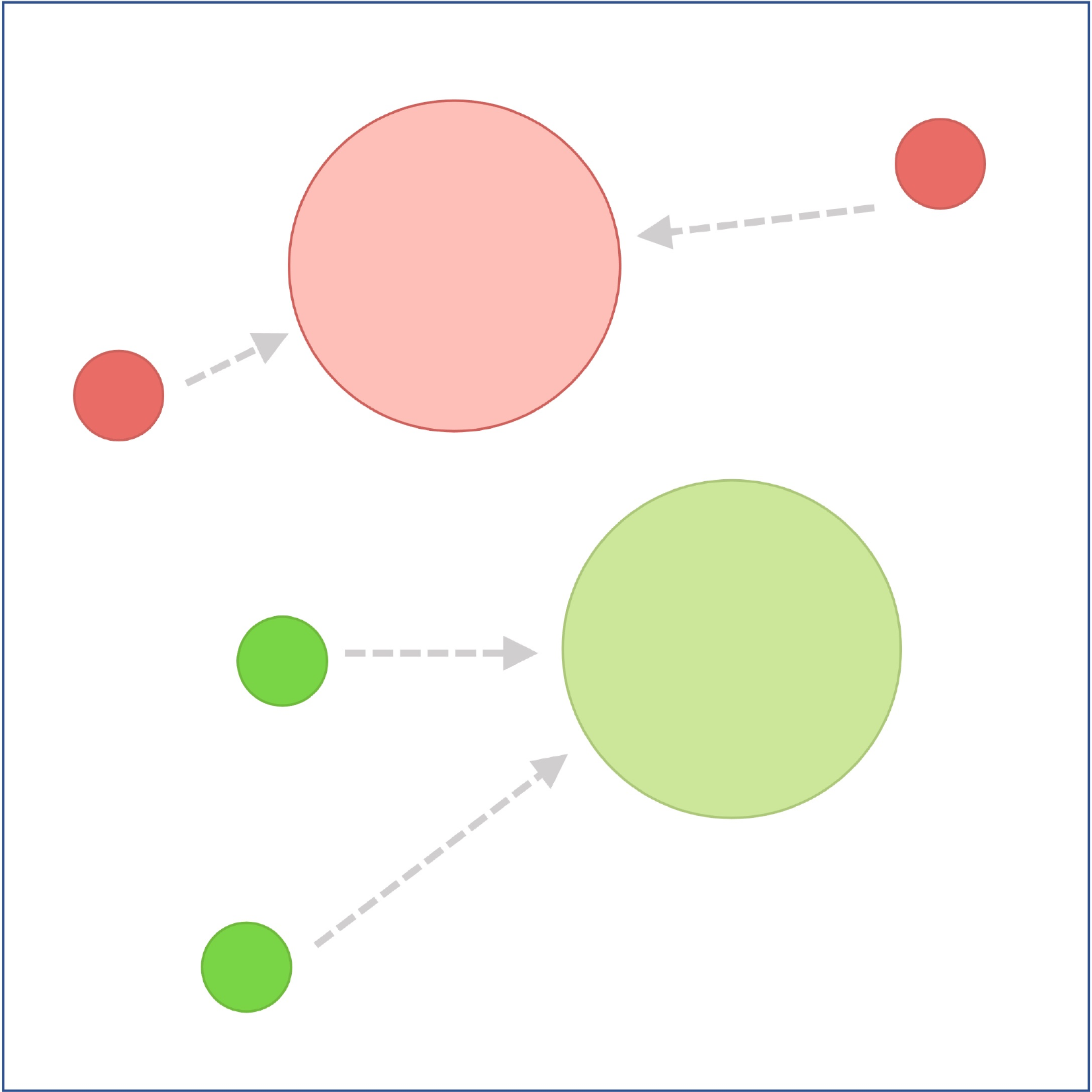}
    \caption{Finding Home.}
    \label{fig:finding_home}
  \end{subfigure}
  \begin{subfigure}{0.4\textwidth}
  \centering
    \includegraphics[width=0.7\linewidth]{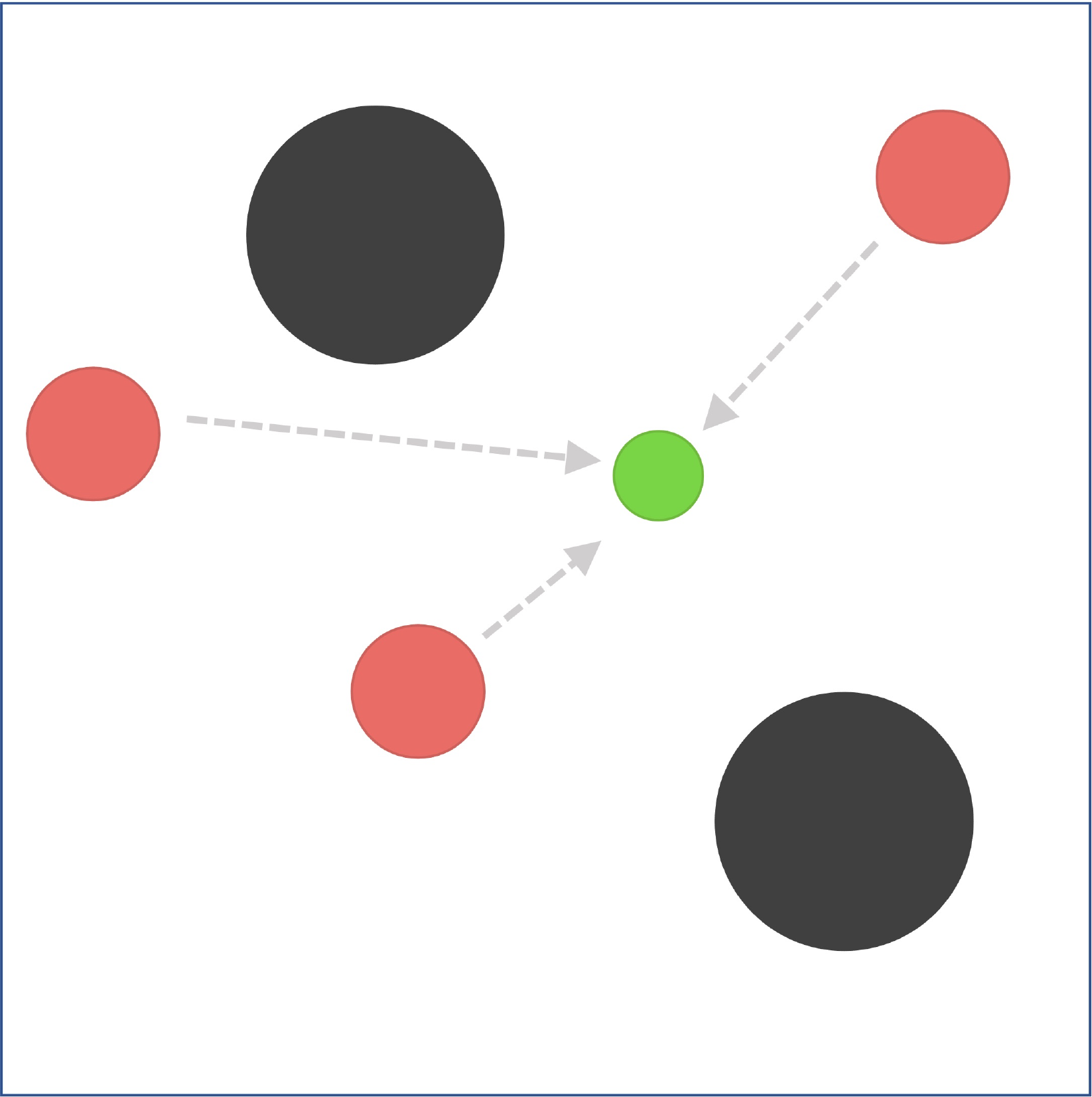}
    \caption{Predator-Prey.}
    \label{fig:predator_prey}
  \end{subfigure}
  \caption{Illustrations of Finding Home and Predator-Prey. In Finding Home, agents of different colors need to enter the landmarks (Homes) with the corresponding colors. In Predator-Prey, predators (red agents) chase the prey (green agent), and the black landmarks are obstacles that impede the way.}
  \label{fig:task_illustration}
\end{figure}

We evaluate our approach in two environments: Finding Home and Predator-Prey. %one is in cooperative setting and the other is in mixed setting. 
The former is built upon the multi-agent particle environment framework\footnote{https://github.com/openai/multiagent-particle-envs}~\cite{mordatch2017emergence,lowe2017multi}, and the latter from the framework is slightly modified to meet our needs. Figure~\ref{fig:task_illustration} illustrates the two tasks. In the experiments, we use discrete action spaces to simplify the control problem, and the actions are moving left, right, up, down, and staying in position. Agents observe the relative positions and velocities of other agents as well as the relative positions of all the landmarks.

% \begin{wrapfigure}[32]{r}{0.35\textwidth}
% \begin{center}
%   \includegraphics[width=0.9\linewidth]{finding_home.png}
%   \caption{Finding Home. Agents of different colors need to enter the large landmarks (homes) with the corresponding colors. }
%   \label{fig:find_home}
%   \smallskip\par
%   \includegraphics[width=0.9\linewidth]{predator-prey.png}
%   \caption{Predator-Prey. Predators (red agents) chase the prey (green agent), the black landmarks are obstacles that impede the way.}
%   \label{fig:predator_prey}
% \end{center}
% \end{wrapfigure}

\paragraph{Finding Home.}
This is a cooperative task where all the agents share the same reward. The environment consists of red agents, green agents, and their homes (landmarks) which are marked with the corresponding colors. 
In this task, the goal of every agent is to learn which home it belongs to and then enters its home while avoiding colliding with others. 
Agents will be punished by the sum of the distances from their homes as well as the number of collisions. 
% The positions of the agents and the homes are randomly initialized. 

\paragraph{Predator-Prey.}
This is a mixed cooperative-competitive environment consisting of predators, their prey, and some landmarks impeding the way. 
The predators work collaboratively to chase the prey and they share the same reward: they will be rewarded if any of them touches the prey, and they will be penalized by the sum of the distances from the prey. 
The prey will be rewarded and penalized oppositely.

%---------------------------------Figure---------------------------------%
%---------------------------------Figure---------------------------------%

\subsection{Model Training and Parameter Settings}
 We use a replay buffer of maximum length 1e6. The target network update rate $\tau$ is set to 7e-5. The discount factor $\gamma$ is set to 0.99. One episode lasts for 25 steps, and the value networks are updated 4 times every 100 steps (4 episodes). We use TD3 to stabilize training, the policies and target networks are updated once every 2 critics updates. When updating the networks, we sample 1024 past experiences from the replay buffer, and we use Adam~\cite{kingma2014adam} optimizer to update the networks. The output dimension of the value network feature extractor is 64 and it is 128 for the policy network feature extractor. The hidden dimensions of all feature extractors are 32, and we use 2 attention heads for the multi-head self-attention. The hidden dimension of each candidate block in the ensemble heads is 64. All the reported models are trained using the best learning rates found by grid search, and all the models are trained on a single NVIDIA TITAN V GPU with 5 different seeds.

\subsection{Effectiveness}
%---------------------------------Figure---------------------------------%
\begin{figure}
\centering
  \begin{subfigure}{0.48\textwidth}
  \centering
    \includegraphics[width=\linewidth]{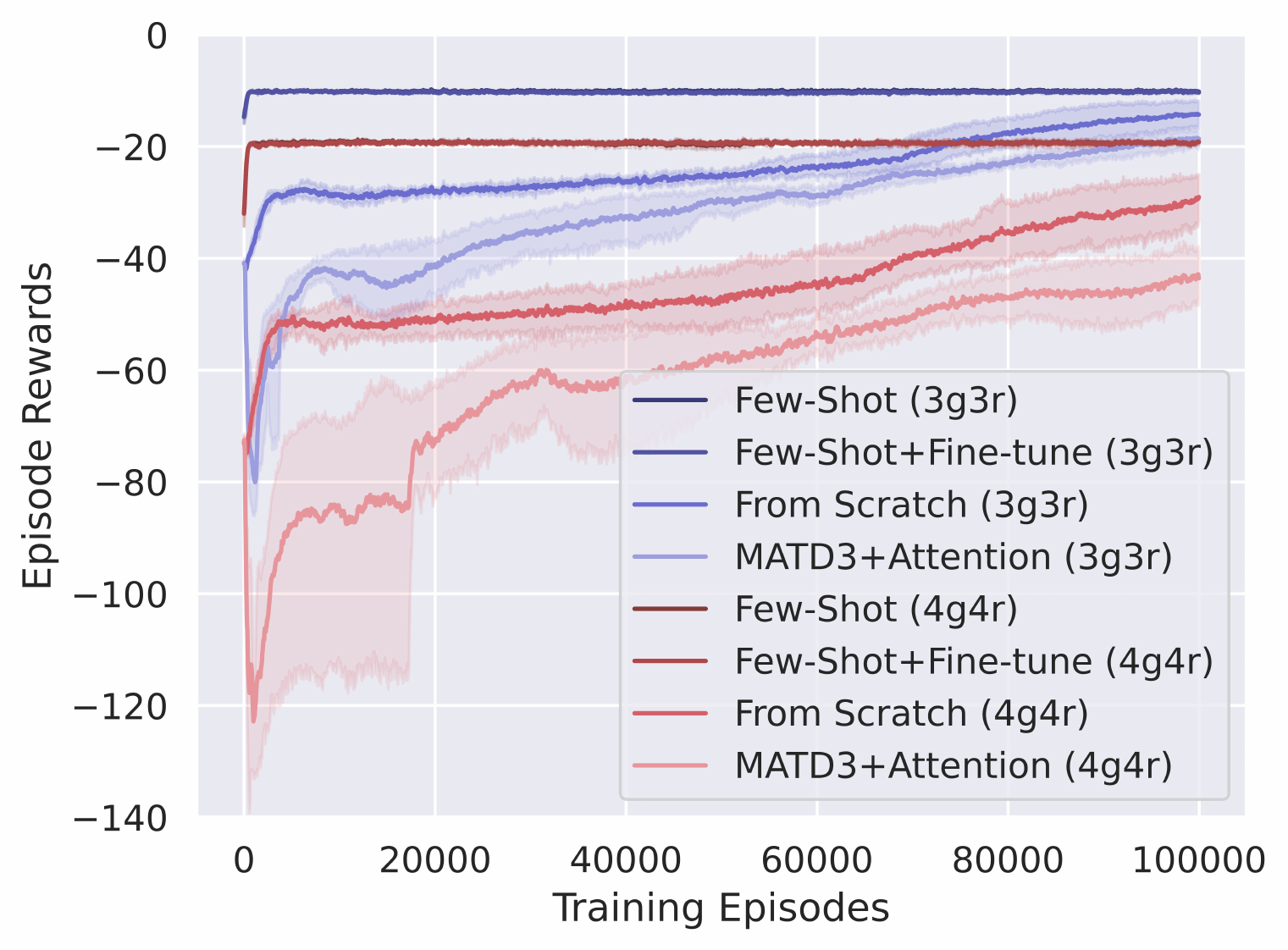}
    \caption{New Agents Participate}
    \label{fig:more_effectiveness}
  \end{subfigure}
  \begin{subfigure}{0.48\textwidth}
  \centering
    \includegraphics[width=\linewidth]{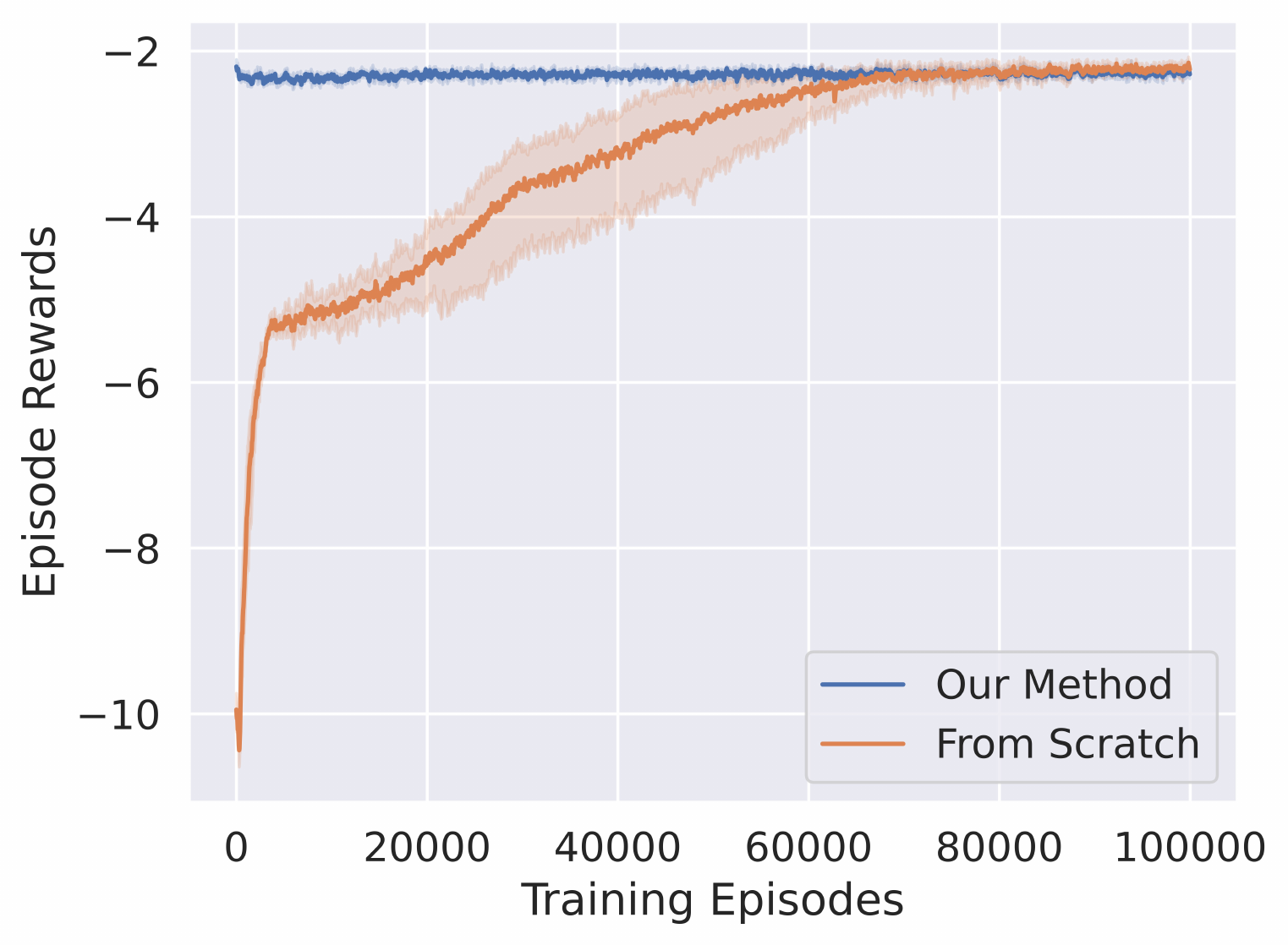}
    \caption{One Agent Drops}
    \label{fig:less_effectiveness}
  \end{subfigure}
  \caption{Average episode rewards on different scenarios after the number of agents changes. (a) shows the rewards of different approaches after more agents join the training. (b) shows the rewards of our method and that of training from scratch on Finding Home after 1 red agent drops from the training. Error bars are 95\% confidence intervals across 5 runs with different seeds. }
  \label{fig:effectiveness}
\end{figure}
%---------------------------------Figure---------------------------------%
We evaluate the effectiveness of our approach in handling dynamic participating agents during training. 
% We will first showcase the robustness of our model when new agents join the training, then we demonstrate that our model can handle dropping agents effortlessly. 
% Since the goal of this work is to avoid re-training an almost properly trained model due to the number of agents changes during training, we first train the models for a certain amount of episodes, then we alter the number of agents to evaluate the effectiveness of our model. 
In the following experiments, the models are initially trained for a certain amount of episodes, then we alter the number of agents to evaluate the effectiveness. Specifically, the model for Finding Home is initially trained for 150k episodes and the model for Predator-Prey is trained for 300k episodes. 
% All the experiments are conducted using 5 different seeds.

\subsubsection{New Agents Join}
We evaluate our method on both tasks.
For Finding Home, our model is initially trained in an environment with 2 green agents and 2 red agents (2g2r). 
Then more agents are added to the training. 
Figure \ref{fig:more_effectiveness} shows the average episode rewards during training after the number of agents is altered to 3 green 3 red (3g3r) and 4 green 4 red (4g4r). 
Here, we compare three different approaches: 1) ``Few-Shot''--- few-shot training as described in Section \ref{sec:few-shot};  2) ``Few-Shot + Fine-Tune''--- few-shot training for 1k episodes then fine-tuning the whole model; 3) ``From Scratch''--- training a new model from scratch. 

As shown in Figure \ref{fig:more_effectiveness}, for both 3g3r and 4g4r, the rewards of the corresponding ``Few-Shot'' and ``Few-Shot + Fine-Tune'' overlap, and these two approaches achieve high rewards at the early stage (at around 1k episodes). This is because the pre-trained shared part of the model is a good start and there are only a few parameters to train. The models achieve strong performance by doing few-shot learning along, and additional fine-tuning does not improve the performance significantly.
On the contrary, in both scenarios, ``From Scratch'' takes significantly more effort: it takes more than 100k episodes to converge on 3g3r, and it takes even longer on 4g4r. In both experiments, our method is more than 100 times faster than training a new model all over again.  
% We also compare the performance of our model with the existing MATD3 algorithm. 
% In order to get fair comparison, we added attention mechanism to MATD3. 
% Here, we compare the performance of the models trained from scratch, since MATD3 is not designed to handle dynamic agents during training. 
% As shown in Figure \ref{fig:more_effectiveness}, compared to MATD3 with attention, our model achieves higher rewards in less episodes. Therefore, our method is comparable or even better than MATD3 in terms of training speed and performance.
% From these two experiments, we can see that, when dealing with newly joined agents, our approach is more than 100x faster than training a new model all over again. 
% Similarly, in Figure \ref{fig:2_more_each_effectiveness}, ``Few-Shot'' and ``Few-Shot + Fine-Tune'' converges at almost the same rewards at around 1000 episodes, and training from scratch needs even more episodes to achieve equivalent performance. 

For Predator-Prey, the model is initially trained in an environment with 3 predator agents and 1 prey agent, then more predators join the training.
% Here, we will showcase the comparison between few-shot learning with fine-tuning and training from scratch. 
Since this is a mixed cooperative-competitive environment, agents' rewards do not provide much information. 
Instead, we let the two groups of agents use different approaches to compare their performance. 
Specifically, we let the predators use the model trained by our approach and the prey use the model trained from scratch, and we count the number of prey touches at different stages of the training. 
Then we switch their approaches and count again.
We evaluate the approaches in different scenarios where 1, 2, and 3 more predators are added to the training.
The results are shown in Table~\ref{tab:predator-prey_compare}.

As shown in Table~\ref{tab:predator-prey_compare}, when predators use policies obtained by our approach, at the early stage of training, the prey is touched frequently, indicating the predators are better at catching the prey than the prey is at escaping. 
It is clear that, at the early stage, our approach outperforms training from scratch by a large margin.
As the training continues, the prey is touched less frequently, this is because the prey models trained from scratch are gradually improved during their training.  
In the opposite scenarios, where the predators use re-trained models and the prey uses models trained by our approach, the results are expected:
at the early training stage, the prey is barely touched, demonstrating the prey is better at escaping than the predators are at catching, the prey is touched more often as the models trained from scratch are improved by training for more episodes.

By evaluating our method in environments of different settings, we demonstrate that our method can effectively handle new participating agents in terms of avoiding re-training models, shortening new agents' adaptation time as well as achieving strong performance. 

%---------------------------------Table---------------------------------%
\begin{table}
    \caption{Average number of prey touches by predators at different training stages on 10 independent runs.}
    \label{tab:predator-prey_compare}
    \begin{center}
    \begin{tabular}{ccrrrr}
    \toprule
        \multirow{2}{*}{Predator $\pi$} & \multirow{2}{*}{Prey \textbf{$\pi$}} & \multirow{2}{2.8cm}{\centering Number of Training Episodes} & \multicolumn{3}{c}{ More Predators}\\
         &  &  & +1 & +2 & +3 \\\toprule
        \multirow{3}{*}{Our Method} & \multirow{3}{*}{From scratch} & 10k & 10.11 & 8.89 & 2.35\\
        & & 50k & 2.01 & 2.09 & 1.15\\ 
        & & 100k & 0.97 & 1.00 & 1.35\\ \midrule
        \multirow{3}{*}{From scratch} & \multirow{3}{*}{Our Method} & 10k & 0.05 & 0.05 & 0.31\\ 
        & & 50k & 0.33 & 0.10 & 0.45\\ 
        & & 100k & 0.31 & 0.42 & 0.32\\ \bottomrule
    \end{tabular}
    \end{center}
\end{table}
%---------------------------------Table---------------------------------%
% %---------------------------------Figure---------------------------------%
% \begin{wrapfigure}[16]{r}{0.5\textwidth}
% \vspace{-1mm}
%   \begin{center}
%     \includegraphics[width=0.45\textwidth]{dynamic_training_1_less_effectiveness.png}
%     \vspace{-1mm}
%   \end{center}
%     \caption{Average episode rewards of our method and training from scratch on Finding Home task after 1 red agent drops from the training.}
%     \label{fig:1_less_effectiveness}
% \vspace{-1mm}
% \end{wrapfigure}
% %---------------------------------Figure---------------------------------%

\subsubsection{Agents Drop}

We also evaluate our method on handling dropping agents during training. For our network structure, dropping agents do not affect the training process of the remaining agents and the model does not need any adaptation, the missing agents are just ignored. Figure \ref{fig:less_effectiveness} shows the average episode rewards on Finding Home after 1 red agent drops from the training.  As shown in Figure \ref{fig:less_effectiveness}, our method does not need any adjustments and achieves high rewards from the beginning, while the new model trained from scratch reaches the same average rewards after 70k episodes.

\subsection{Comparison with Existing Method}
We also compare the performance of our model with the existing MATD3 model. 
% Here, we compare the performance of the models trained from scratch, since MATD3 is not designed to handle dynamic agents during training. 
Since MATD3 is not designed to handle dynamic agents during training, it must be re-trained when the number of agents changes.
To get a fair comparison, we add self-attention to MATD3. 
As shown in Figure \ref{fig:more_effectiveness}, compared to MATD3 with attention (``MATD3 + Attention''), our model (``From Scratch'') achieves higher rewards in fewer episodes. Therefore, our model is comparable to or better than MATD3 in terms of training speed and performance. For new agents adaptation, our model trained by few-shot learning is more than 100 times faster than ``MATD3 + Attention''.

\subsection{Ablation Study}
\begin{figure}
\centering
  \begin{subfigure}{0.48\textwidth}
  \centering
    \includegraphics[width=\linewidth]{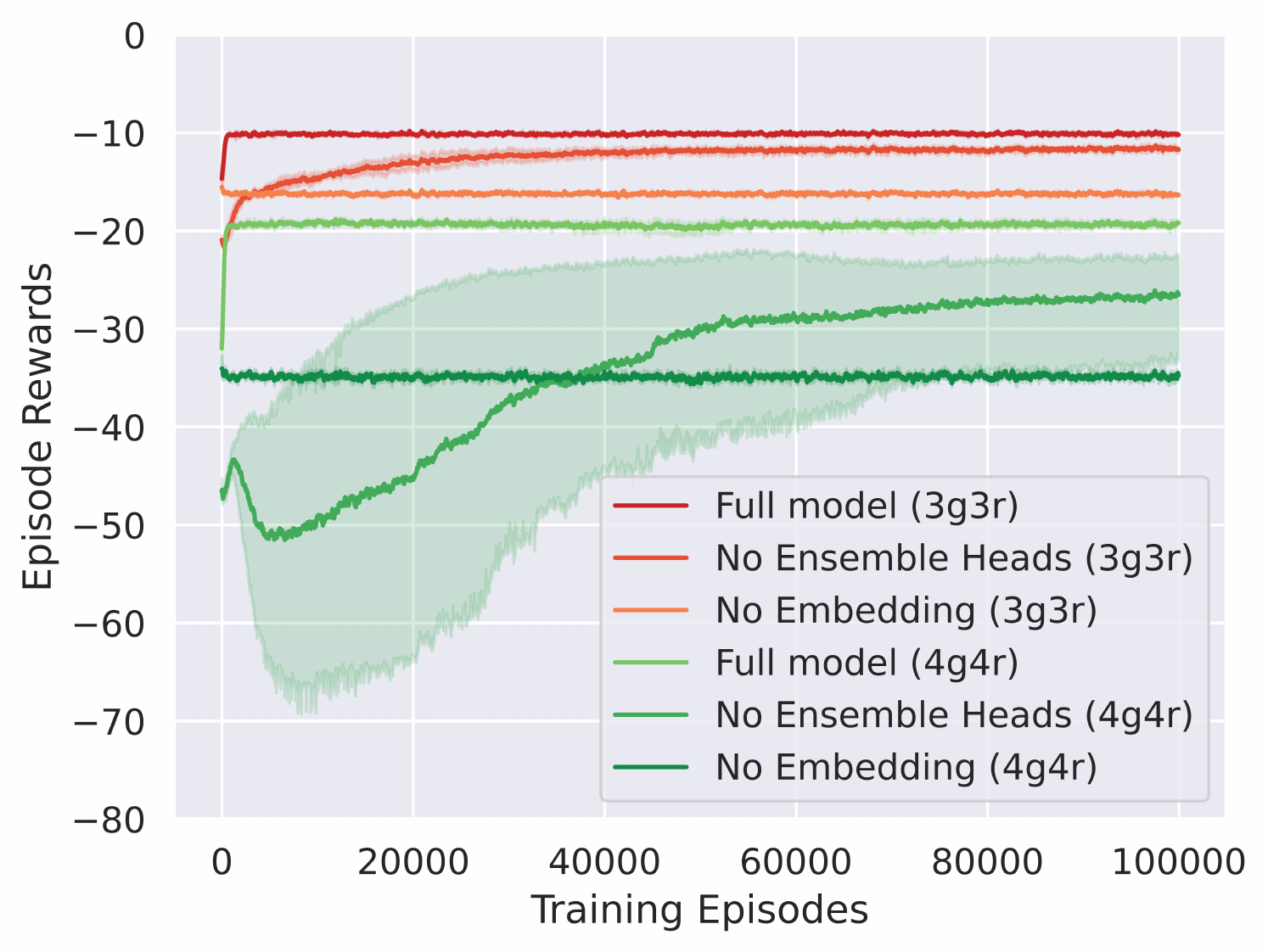}
    \caption{Ablation Study}
    \label{fig:ablation}
  \end{subfigure}
  \begin{subfigure}{0.48\textwidth}
  \centering
    \includegraphics[width=\linewidth]{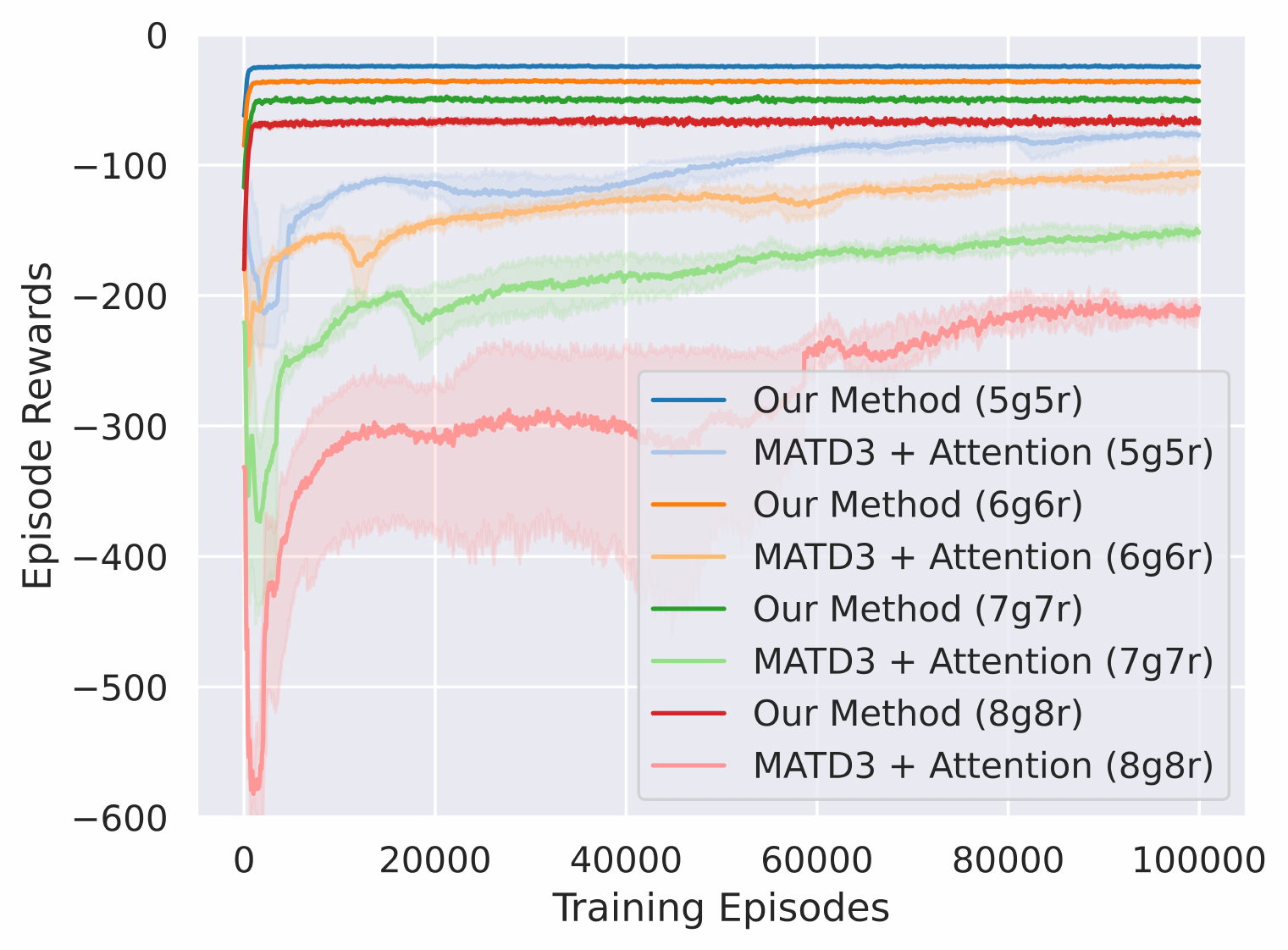}
    \caption{Scalability}
    \label{fig:scalability}
  \end{subfigure}
  \caption{(a) shows the average episode rewards of the full model (``Full Model''), the model with no agent embeddings (``No Embedding''), and the model with independent MLPs heads (``No Ensemble Heads'') after 2g2r is altered to 3g3r and 4g4r; (b) shows the scalability of our method and that of MATD3 with attention, the environment is initially 2g2r, and it is increased to 5g5r all the way to 8g8r. Error bars are 95\% confidence intervals across 5 runs with different seeds. }
  \label{fig:ablation_and_scalability}
\end{figure}

We conduct ablation studies to analyze the importance of two essential components: the ensemble heads and the agent ID embeddings. 
We compare the training speed and average episode rewards of three models: 1) the full model; 2) the model with no agent embeddings but has ensemble heads; 3) the model with no ensemble heads but has agent embedding, it uses independent MLPs as its heads. 
We perform the ablation studies on Finding Home. The model is initially trained on 2g2r, then the number of agents is altered to 3g3r and 4g4r.
% For model with no agent embeddings, we train its ensemble heads during few-shot learning, and for model with no ensemble heads, we train its MLPs heads as well as its agent embeddings. 
For all models, only the independent components are trained, and the shared components are fixed.
The results are shown in Figure \ref{fig:ablation}. 
The average rewards of the "No Embedding" do not increase during its training, which shows that agent ID embeddings are essential to our model.
Without shared ensemble heads, it takes ``No Ensemble Heads'' significantly longer to get properly trained since the MLP heads have more parameters and need more episodes to train.

\subsection{Scalability}

% As the number of new participating agents grows, training from scratch takes more and more resources and time, in contrast, the adaptation using our approach is still quick and easy.
Figure~\ref{fig:scalability} demonstrates the comparison of the scalability of our method and that of MATD3 with attention on Finding Home. The initial environment is 2g2r, and then more agents are added during the training. 
% For our method, the training of the new agents can be done within a few shots, while training MATD3 with attention from scratch takes increasingly more episodes.
The figure shows that, as the number of new participating agents increases, the gap between our method and MATD3 with attention becomes larger; training MATD3 with attention becomes slower and it takes increasingly more episodes, while our method can still train the new agents in a few shots. In these experiments, our method demonstrates strong scalability.

%our approach finishes the training more than 100x faster than training from scratch, and the speedup is increasing along with the growth.

\subsection{Visualizing Selectors}
\begin{figure}
\centering
  \begin{subfigure}{0.48\textwidth}
  \centering
    \includegraphics[width=0.9\linewidth]{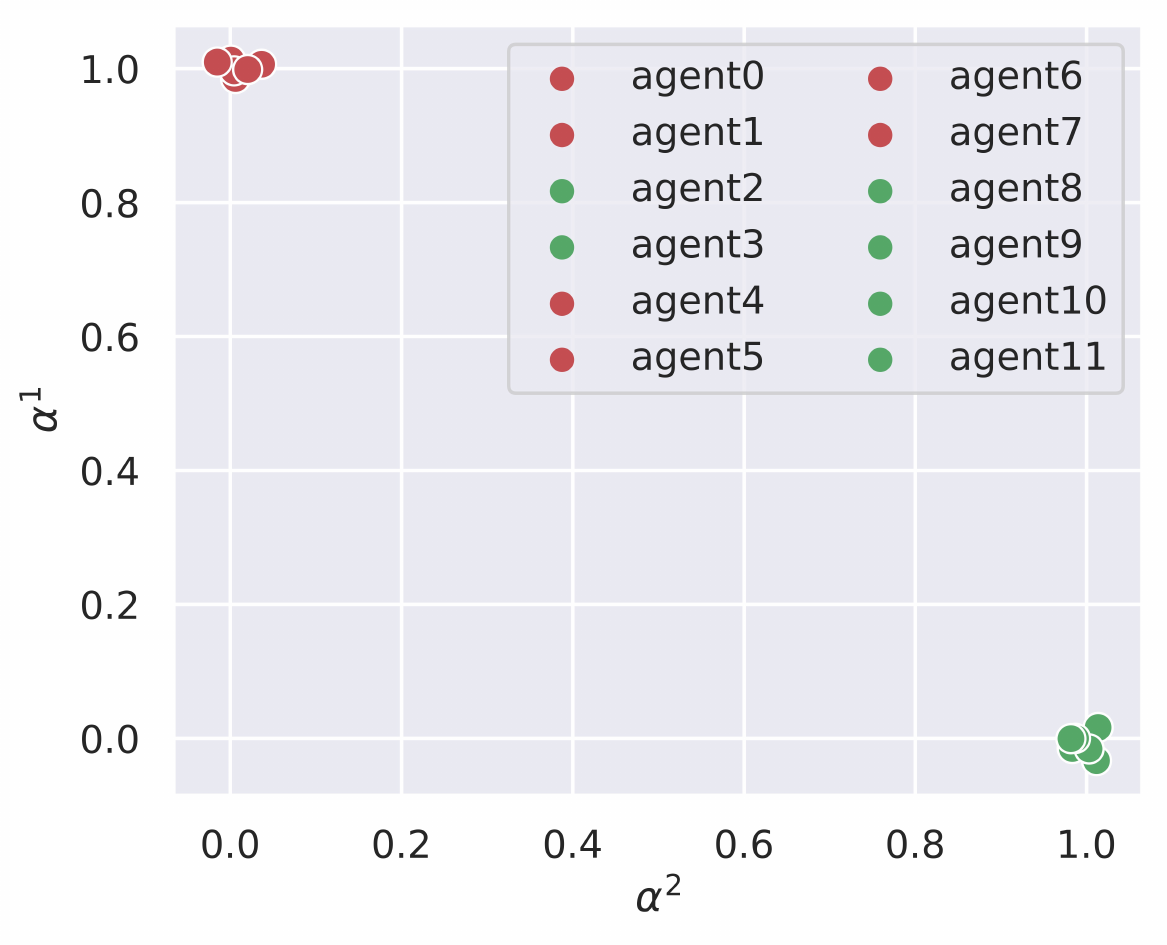}
    \caption{Value Candidate Selectors}
    \label{fig:value_selector}
  \end{subfigure}
  \begin{subfigure}{0.48\textwidth}
  \centering
    \includegraphics[width=0.9\linewidth]{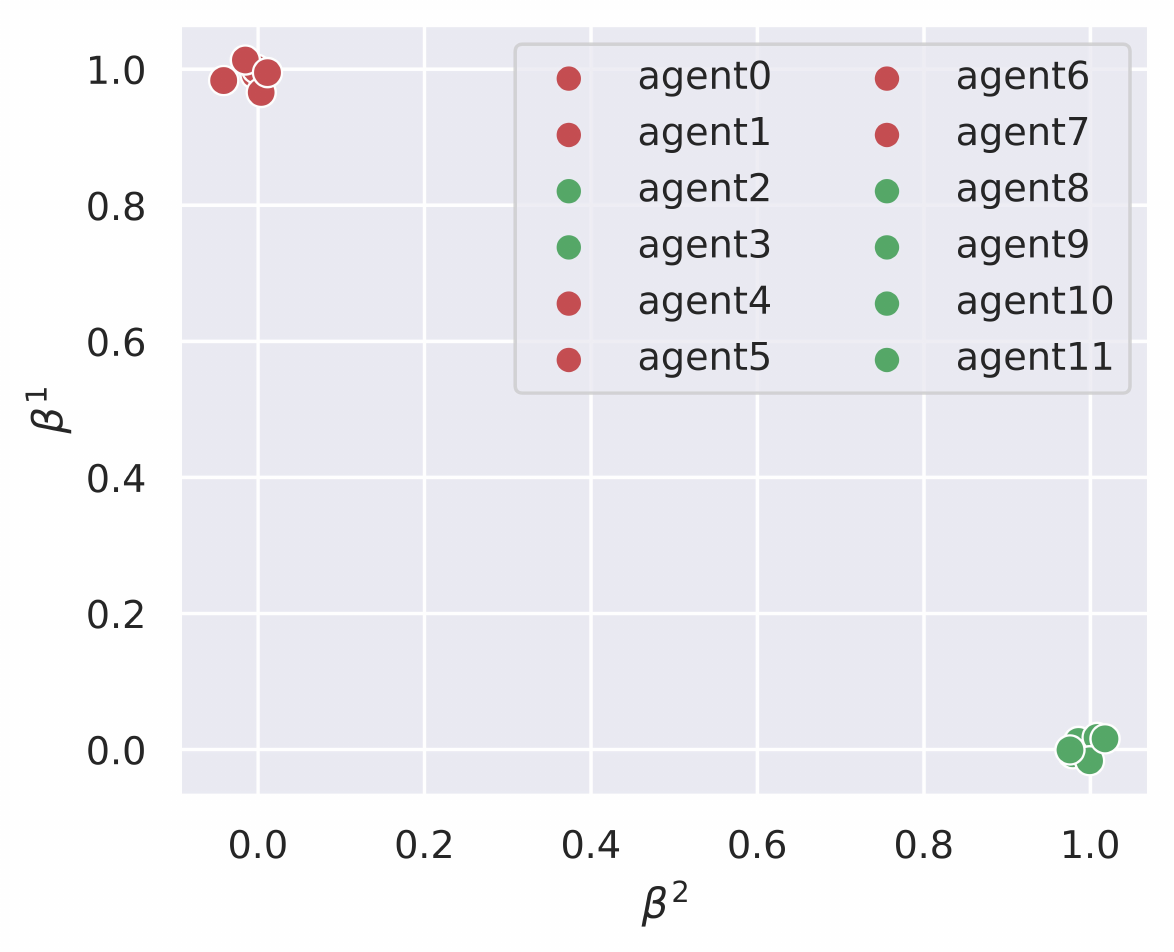}
    \caption{Policy Candidate Selectors}
    \label{fig:policy_selector}
  \end{subfigure}
  \caption{The values (after softmax operation) of the selectors of all agents. (a) shows the selectors of value candidate blocks. (b) shows the selectors of policy candidate blocks. There are 2 types of agents in Finding Home, thus the selectors are 2-dimensional. Agent 0 to agent 3 are original agents, and the rest are new agents. The legend colors reflect the true colors of the agents. The points are jittered with random noises to increase their visibility in the plot. }
  \label{fig:selectors}
\end{figure}

After new agents join the training, the selectors in their value networks and policy networks should be properly trained by few-shot learning. Figure~\ref{fig:selectors} visualizes the selectors of a model which is few-shot trained on Finding Home. The original environment is 2g2r and it is altered to 6g6r. 
As shown in the figures, after the few-shot learning, the selectors of different types of agents are distinguishable, and agents of the same type prefer the same candidate block.

\section{Conclusion} \label{sec:conclusion}

We proposed a novel neural network architecture and a few-shot learning algorithm for deep MARL.
Our approach allows the number of agents to vary during training.
When new agents participate halfway during training, the new agents' policy networks and value networks can get trained in a few shots.
If some agents drop from the training, the remaining agents can still be trained without any adjustment.
Our approach is applicable to cooperative, competitive, and mixed environments.
Our experiments demonstrated the effectiveness and scalability of our work.
The limitation of our work is that a newly joined agent must have sufficient similarity with some of the existing agents in the system.

% We show the effectiveness of our method, including how much our method can speed up the adaptation process when new agents join as well as how competitive the performance can be after adapting in a few shots. We demonstrate the importance of the key architecture components. Besides, we demonstrate the scalability of our method. This work has its limitation, we only evaluate the effectiveness on seen agent types, in the future, we intend to evaluate it on unseen types or design a model with better transferring capabilities to unseen types.

\newpage

\begin{small}
% 	\bibliography{rl}
	\bibliographystyle{plain}

\end{small}

\end{document}

%% file: newcommands.tex
\newcommand{\mylabel}[2]{#2\def\@currentlabel{#2}\label{#1}}
\makeatother

\usepackage{enumitem}
\setlist[enumerate,1]{leftmargin=*,wide=0em, label = {\bfseries \roman*.}}
\setlist[itemize,1]{leftmargin=*,wide=0em, itemsep=0pt,topsep=0pt}
\def\AM{{\mathcal A}}

\def\OM{{\mathcal O}}
\def\SM{{\mathcal S}}
\def\TM{{\mathcal T}}

\def\RB{{\mathbb R}}

\def\pii{\mbox{\boldmath$\pi$\unboldmath}}

%% file: rl.bbl
\begin{thebibliography}{10}

\bibitem{ackermann2019reducing}
Johannes Ackermann, Volker Gabler, Takayuki Osa, and Masashi Sugiyama.
\newblock Reducing overestimation bias in multi-agent domains using double
  centralized critics.
\newblock {\em arXiv preprint arXiv:1910.01465}, 2019.

\bibitem{bahdanau2014neural}
Dzmitry Bahdanau, Kyunghyun Cho, and Yoshua Bengio.
\newblock Neural machine translation by jointly learning to align and
  translate.
\newblock {\em arXiv preprint arXiv:1409.0473}, 2014.

\bibitem{barto1983neuronlike}
Andrew~G Barto, Richard~S Sutton, and Charles~W Anderson.
\newblock Neuronlike adaptive elements that can solve difficult learning
  control problems.
\newblock {\em IEEE transactions on systems, man, and cybernetics},
  (5):834\textasciitilde846, 1983.

\bibitem{cheng2016long}
Jianpeng Cheng, Li~Dong, and Mirella Lapata.
\newblock Long short-term memory-networks for machine reading.
\newblock {\em arXiv preprint arXiv:1601.06733}, 2016.

\bibitem{foerster2018counterfactual}
Jakob Foerster, Gregory Farquhar, Triantafyllos Afouras, Nantas Nardelli, and
  Shimon Whiteson.
\newblock Counterfactual multi-agent policy gradients.
\newblock In {\em AAAI Conference on Artificial Intelligence (AAAI)}, 2018.

\bibitem{foerster2017stabilising}
Jakob Foerster, Nantas Nardelli, Gregory Farquhar, Triantafyllos Afouras,
  Philip H.~S. Torr, Pushmeet Kohli, and Shimon Whiteson.
\newblock Stabilising experience replay for deep multi-agent reinforcement
  learning, 2017.

\bibitem{fujimoto2018addressing}
Scott Fujimoto, Herke Hoof, and David Meger.
\newblock Addressing function approximation error in actor-critic methods.
\newblock In {\em International conference on machine learning}, pages
  1587--1596. PMLR, 2018.

\bibitem{hu2020updet}
Siyi Hu, Fengda Zhu, Xiaojun Chang, and Xiaodan Liang.
\newblock Updet: Universal multi-agent rl via policy decoupling with
  transformers.
\newblock In {\em International Conference on Learning Representations}, 2020.

\bibitem{iqbal2021randomized}
Shariq Iqbal, Christian A~Schroeder De~Witt, Bei Peng, Wendelin B{\"o}hmer,
  Shimon Whiteson, and Fei Sha.
\newblock Randomized entity-wise factorization for multi-agent reinforcement
  learning.
\newblock In {\em International Conference on Machine Learning}, pages
  4596--4606. PMLR, 2021.

\bibitem{iqbal2019actor}
Shariq Iqbal and Fei Sha.
\newblock Actor-attention-critic for multi-agent reinforcement learning.
\newblock In {\em International Conference on Machine Learning}, pages
  2961--2970. PMLR, 2019.

\bibitem{kingma2014adam}
Diederik~P Kingma and Jimmy Ba.
\newblock Adam: A method for stochastic optimization.
\newblock {\em arXiv preprint arXiv:1412.6980}, 2014.

\bibitem{lazaridou2016multiagent}
Angeliki Lazaridou, Alexander Peysakhovich, and Marco Baroni.
\newblock Multi-agent cooperation and the emergence of (natural) language,
  2016.

\bibitem{littman1994markov}
Michael~L Littman.
\newblock Markov games as a framework for multi-agent reinforcement learning.
\newblock In {\em Machine learning proceedings 1994}, pages 157--163. Elsevier,
  1994.

\bibitem{liu2021coach}
Bo~Liu, Qiang Liu, Peter Stone, Animesh Garg, Yuke Zhu, and Anima Anandkumar.
\newblock Coach-player multi-agent reinforcement learning for dynamic team
  composition.
\newblock In {\em International Conference on Machine Learning}, pages
  6860--6870. PMLR, 2021.

\bibitem{lowe2017multi}
Ryan Lowe, Yi~I Wu, Aviv Tamar, Jean Harb, OpenAI~Pieter Abbeel, and Igor
  Mordatch.
\newblock Multi-agent actor-critic for mixed cooperative-competitive
  environments.
\newblock In {\em Advances in Neural Information Processing Systems (NIPS)},
  2017.

\bibitem{mordatch2017emergence}
Igor Mordatch and Pieter Abbeel.
\newblock Emergence of grounded compositional language in multi-agent
  populations.
\newblock {\em arXiv preprint arXiv:1703.04908}, 2017.

\bibitem{pmlr-v80-rashid18aQMIX}
Tabish Rashid, Mikayel Samvelyan, Christian Schroeder, Gregory Farquhar, Jakob
  Foerster, and Shimon Whiteson.
\newblock {QMIX}: Monotonic value function factorisation for deep multi-agent
  reinforcement learning.
\newblock In {\em Proceedings of the 35th International Conference on Machine
  Learning}, pages 4295--4304, 2018.

\bibitem{sukhbaatar2016learning}
Sainbayar Sukhbaatar, Arthur Szlam, and Rob Fergus.
\newblock Learning multiagent communication with backpropagation, 2016.

\bibitem{Tampuu_2017DQN_IQL}
Ardi Tampuu, Tambet Matiisen, Dorian Kodelja, Ilya Kuzovkin, Kristjan Korjus,
  Juhan Aru, Jaan Aru, and Raul Vicente.
\newblock Multiagent cooperation and competition with deep reinforcement
  learning.
\newblock {\em PLOS ONE}, 12(4):e0172395, Apr 2017.

\bibitem{tan1993multi}
Ming Tan.
\newblock Multi-agent reinforcement learning: Independent versus cooperative
  agents.
\newblock In Paul~E. Utgoff, editor, {\em 10th International Conference on
  Machine Learning, Amherst, MA, USA, 1993}, Burlington, Massachusetts, 1993.
  Morgan Kaufmann.

\bibitem{vaswani2017attention}
Ashish Vaswani, Noam Shazeer, Niki Parmar, Jakob Uszkoreit, Llion Jones,
  Aidan~N Gomez, {\L}ukasz Kaiser, and Illia Polosukhin.
\newblock Attention is all you need.
\newblock {\em Advances in neural information processing systems}, 30, 2017.

\end{thebibliography}
